\documentclass[11pt,a4paper]{article}
\usepackage[hyperref]{emnlp-ijcnlp-2019-latex/emnlp-ijcnlp-2019}
\usepackage{times}
\usepackage{latexsym}
\pdfoutput=1

\usepackage{amsmath,amssymb,amsfonts}
\usepackage{url,enumitem}
\usepackage{gb4e-safe}
\usepackage{booktabs}
\usepackage{xspace}

\newcommand{\wnut}{\textsc{wnut}\xspace}
\newcommand{\conll}{\textsc{CoNLL}\xspace}

\usepackage{notation}

\aclfinalcopy 
\def\aclpaperid{2686} 

\title{Unsupervised Domain Adaptation of Contextualized Embeddings for Sequence Labeling}

\author{Xiaochuang Han \and Jacob Eisenstein\\
Georgia Institute of Technology\thanks{~~XH is now at Carnegie Mellon University and JE is now at Google Research. Some of the work was performed while JE was visiting Facebook AI Research.}\\
  \texttt{xiaochuang.han@gmail.com, me@jacob-eisenstein.com} \\}

\date{}

\begin{document}
\maketitle
\begin{abstract}
  Contextualized word embeddings such as ELMo and BERT provide a foundation for strong performance across a wide range of natural language processing tasks by pretraining on large corpora of unlabeled text.
However, the applicability of this approach is unknown when the target domain varies substantially from the pretraining corpus.
We are specifically interested in the scenario in which labeled data is available in only a canonical source domain such as newstext, and the target domain is distinct from both the labeled and pretraining texts.
To address this scenario, we propose \emph{domain-adaptive fine-tuning}, in which the contextualized embeddings are adapted by masked language modeling on text from the target domain.
We test this approach on sequence labeling in two challenging domains: Early Modern English and Twitter. Both domains differ substantially from existing pretraining corpora, and domain-adaptive fine-tuning yields substantial improvements over strong BERT baselines, with particularly impressive results on out-of-vocabulary words.
We conclude that domain-adaptive fine-tuning offers a simple and effective approach for the unsupervised adaptation of sequence labeling to difficult new domains.\footnote{Trained models for Early Modern English and Twitter are available at \url{https://github.com/xhan77/AdaptaBERT}}

\end{abstract}

\section{Introduction}
Contextualized word embeddings are becoming a ubiquitous component of natural language processing~\cite{dai2015semi,devlin2018bert,howard2018universal,radford2018improving,peters2018deep}.
Pretrained contextualized word embeddings can be used as feature for downstream tasks; alternatively, the contextualized word embedding module can be incorporated into an end-to-end system, allowing the embeddings to be fine-tuned from task-specific labeled data.
In either case, a primary benefit of contextualized word embeddings is that they seed the learner with distributional information from large unlabeled datasets.

However, the texts used to build pretrained contextualized word embedding models are drawn from a narrow set of domains:
\begin{itemize}[itemsep=0pt]
\item \textbf{Wikipedia} in BERT~\cite{devlin2018bert} and ULMFiT~\cite{howard2018universal};
\item \textbf{Newstext}~\cite{chelba2013one} in ELMo~\cite{peters2018deep};
\item \textbf{BooksCorpus}~\cite{zhu2015aligning} in BERT~\cite{devlin2018bert} and GPT~\cite{radford2018improving}.
\end{itemize}
All three corpora consist exclusively of text written since the late 20th century; furthermore, Wikipedia and newstext are subject to restrictive stylistic constraints~\cite{bryant2005becoming}.\footnote{While it might be desirable to completely retrain contextualized word embedding models in the target domain~\cite[e.g.,][]{lee2019biobert}, this requires data and computational resources that are often unavailable.}
It is therefore crucial to determine whether these pretrained models are transferable to texts from other periods or other stylistic traditions, such as historical documents, technical research papers, and social media.

This paper offers a first empirical investigation of the applicability of domain adaptation to pretrained contextualized word embeddings. As a case study, we focus on sequence labeling in historical texts.
Historical texts are of increasing interest in the computational social sciences and digital humanities, offering insights on patterns of language change~\cite{hilpert2016quantitative}, social norms~\cite{garg2018word}, and the history of ideas and culture~\cite{michel2011quantitative}.
Syntactic analysis can play an important role: researchers have used part-of-speech tagging to identify syntactic changes~\cite{degaetano2018stylistic} and dependency parsing to quantify gender-based patterns of adjectival modification and possession in classic literary texts~\cite{vuillemot2009what,muralidharan2013supporting}.
But despite the appeal of using NLP in historical linguistics and literary analysis, there is relatively little research on how performance is impacted by diachronic transfer.
Indeed, the evidence that does exist suggests that accuracy degrades significantly, especially if steps are not taken to adapt:
for example, \newcite{yang2015unsupervised} compare the accuracy of tagging 18th century and 16th century Portuguese text (using a model trained on 19th century text), and find that the error rate is twice as high for the older text.

We evaluate the impact of diachronic transfer on a contemporary pretrained part-of-speech tagging system.
First, we show that a BERT-based part-of-speech tagger outperforms the state-of-the-art unsupervised domain adaptation method~\cite{yang2016part}, without taking any explicit steps to adapt to the target domain of Early Modern English.
Next, we propose a simple unsupervised domain-adaptive fine-tuning step, using a masked language modeling objective over unlabeled text in the target domain.
This yields significant further improvements, with more than 50\% error reduction on out-of-vocabulary words. We also present a secondary evaluation on named entity recognition in contemporary social media~\cite{strauss2016results}. This evaluation yields similar results: a direct application of BERT outperforms most of the systems from the 2016 \wnut shared task, and domain-adaptive fine-tuning yields further improvements. In both Early Modern English and social media, domain-adaptive fine-tuning does not decrease performance in the original source domain; the adapted tagger performs well in both settings. In contrast, fine-tuning BERT on \emph{labeled} data in the target domain causes catastrophic forgetting~\cite{mccloskey1989catastrophic}, with a significant deterioration in tagging performance in the source domain.

\section{Tagging Historical Texts}
Before describing our modeling approach, we highlight some of the unique aspects of tagging historical texts, focusing on the target dataset of the Penn-Helsinki Corpus of Early Modern English~\cite{kroch2004penn}.

\subsection{Early Modern English}
Early Modern English (EME) refers to the dominant language spoken in England during the period spanning the 15th-17th centuries, which includes the time of Shakespeare.
While the English of this period is more comprehensible to contemporary readers than the Middle English that immediately preceded it, EME nonetheless differs from the English of today in a number of respects.

\paragraph{Orthography.} A particularly salient aspect of EME is the variability of spelling and other orthographic conventions~\cite{baron2008vard2}. For example:
\begin{exe}
  \ex If this marsch waulle (marsh wall) were not kept, and the canales of eche partes of Sowey river kept from abundance of wedes, al the plaine marsch ground at sodaine raynes (sudden rains) wold be overflowen, and the profite of the meade lost.
\end{exe}
While these differences are not too difficult for fluent human readers of English, they affect a large number of tokens, resulting in a substantial increase in the out-of-vocabulary rate~\cite{yang2016part}.
Some of the spelling differences are purely typographical, such as the substitution of \example{v} for \example{u} in words like \example{vnto}, and the substitution of \example{y} for \example{i} in words like \example{hym}. These are common sources of errors for baseline models. Another source of out-of-vocabulary words is the addition of a silent \example{e} to the end of many words. This generally did not cause errors for wordpiece-based models (like BERT), perhaps because the final `e' is segmented as a separate token, which does not receive a tag.
Capitalization is also used inconsistently, making it difficult to distinguish proper and common nouns, as in the following examples:
\begin{exe}
\ex And that those \textbf{Writs} which shall be awarded and directed for returning of \textbf{Juryes} \ldots
\ex \ldots shall not then have \textbf{Twenty} pounds or \textbf{Eight} pounds respectively \ldots
\end{exe}

\paragraph{Morphosyntax.} Aside from orthography, EME is fairly similar to contemporary English, with a few notable exceptions. EME includes several inflections that are rare or nonexistent today, such as the \example{-th} suffix for third person singular conjugation, as in \example{hath} (has) and \example{doth} (does).
Another difference is in the system of second-person pronouns. EME includes the informal second person \example{thou} with declensions \example{thee, thine, thy}, and the plural second-person pronoun \example{ye}. These pronouns are significant sources of errors for baseline models: for example, a BERT-based tagger makes 216 errors on 272 occurrences of the pronoun \example{thee}.

\subsection{Part-of-Speech Tags in the Penn Parsed Corpora of Historical English}
The Penn Parsed Corpora of Historical English (PPCHE) include part-of-speech annotations for texts from several historical periods~\cite{kroch2004penn}.
We focus on the corpus covering Early Modern English, which we refer to as \ppceme.
As discussed in \autoref{sec:rw}, prior work has generally treated tagging the \ppceme as a problem of \emph{domain adaptation}, with a post-processing stage to map deterministically between the tagsets.
Specifically, we train on the Penn Treebank (PTB) corpus of 20th century English~\cite{marcus1993building}, and then evaluate on the \ppceme test set, using a mapping between the PPCHE and PTB tagsets defined by \newcite{moon2007part}.

Unfortunately, there are some fundamental differences in the approaches to verbs taken by each tagset. Unlike the PTB, the PPCHE has distinct tags for the modal verbs \example{have}, \example{do}, and \example{be} (and their infections); unlike the PPCHE, the PTB has distinct tags for third-person singular present indicative (\ptbtag{vbz}) and other present indicative verbs (\ptbtag{vbp}).
\citeauthor{moon2007part} map only to \ptbtag{vbp}, and \citeauthor{yang2016part} report an error when \ptbtag{vbz} is predicted, even though the corresponding PPCHE tag would be identical in both
cases.
We avoid this issue by focusing most of our evaluations on a coarse-grained version of the PTB tagset, described in \autoref{sec:eval-tagset}.

\section{Adaptation with Contextualized Embeddings}
\label{sec:model}
\label{sec:model-fine-tune}
The problem of processing historical English can be treated as one of unsupervised domain adaptation.
Specifically, we assume that labeled data is available only in the ``source domain'' of contemporary modern English, and adapt to the ``target domain'' of historical text by using unlabeled data.
We now explain how contextualized word embeddings can be applied in this setting.

Contextualized word embeddings provide a generic framework for semi-supervised learning across a range of natural language processing tasks, including sequence labeling tasks like part-of-speech tagging and named entity recognition~\cite{peters2018deep,devlin2018bert}. Given a sequence of tokens $w_1, w_2, \ldots, w_T$, these methods return a sequence of vector embeddings $\vx_1, \vx_2, \ldots, \vx_T$. The embeddings are \emph{contextualized} in the sense that they reflect not only each token but also the context in which each token appears. The embedding function is trained either from a language modeling task~\cite{peters2018deep} or a related task of recovering masked tokens~\cite{devlin2018bert}; these methods can be seen as performing semi-supervised learning, because they benefit from large amounts of unlabeled data.

\subsection{Task-specific fine-tuning}
Contextualized embeddings are powerful features for a wide range of downstream tasks. Of particular relevance for our work, \newcite{devlin2018bert} show that a state-of-the-art named entity recognition system can be constructed by simply feeding the contextualized embeddings into a linear classification layer. The log probability can then be computed by the log softmax,
\begin{equation}
  \log p(y_t \mid \vec{w}_{1:T}) = \beta_{y_t} \cdot \vx_t - \log \sum_{y \in \set{Y}} \exp \left( \beta_{y} \cdot \vx_t \right),
  \label{eq:conditional-ll}
\end{equation}
where the contextualized embedding $\vx_t$ captures information from the entire sequence $\vec{w}_{1:T} = (w_1, w_2, \ldots, w_T)$, and $\beta_y$ is a vector of weights for each tag $y$.

To fine-tune the contextualized word embeddings, the model is trained by minimizing the negative conditional log-likelihood of the labeled data.
This involves backpropagating from the tagging loss into the network that computes contextualized word embeddings.
We refer to this procedure as \emph{task-tuning}.

To borrow from the terminology of domain adaptation~\cite{daume2006domain}, a \emph{direct transfer} of contextualized word embeddings to the problem of tagging historical text works as follows:
\begin{enumerate}
\item Fine-tune BERT for the part-of-speech tagging task, using the Penn Treebank (PTB) corpus of 20th century English;
\item Apply BERT and the learned tagger to the test set of the Penn Parsed Corpus of Early Modern English (\ppceme).
\end{enumerate}
We evaluate this approach in \autoref{sec:results}.

\subsection{Domain-adaptive fine-tuning}
When the target domain differs substantially from the pretraining corpus, the contextualized word embeddings may be ineffective for the tagging task. This risk is particularly serious in unsupervised domain adaptation, because the labeled data may also differ substantially from the target text. In this case, task-specific fine-tuning may help adapt the contextualized embeddings to the labeling task, but not to the domain. To address this issue, we propose the \emph{AdaptaBERT} model for unsupervised domain adaptation, which adds an additional fine-tuning objective: masked language modeling in the target domain. Specifically, we apply a simple two-step approach:
\begin{enumerate}
\item
  \textbf{Domain tuning}. In the first step, we fine-tune the contextualized word embeddings by backpropagating from the BERT objective, which is to maximize the log-probability of randomly masked tokens.

  We apply this training procedure to a dataset that includes all available target domain data, and an equal amount of unlabeled data in the source domain.\footnote{If the source domain dataset is smaller than the target domain data, then all of the unlabeled source domain data is included.}
  We create ten random maskings of each instance; in each masking, 15\% of the tokens are randomly masked out, following the original BERT training procedure.
  We then perform three training iterations over this masked data.

\item
\textbf{Task tuning}. In the second step, we fine-tune the contextualized word embeddings and learn the prediction model by backpropagating from the labeling objective on the source domain labeled data (\autoref{eq:conditional-ll}). 
This step fine-tunes the contextualized embeddings for the desired labeling task.
\end{enumerate}
Attempts to interleave these two steps did not yield significant improvements in performance. While \newcite{peters-etal-2019-tune} report good results on named entity recognition without task tuning, we found this step to be essential in our transfer applications.

\begin{table}
    \centering
    \begin{tabular}{@{}lp{1in}p{1in}@{}}
    \toprule
     & Domain tuning & Task tuning\\
      \midrule
      Prediction & masked tokens & tags\\
      Data & source + target & source\\
      \bottomrule
    \end{tabular}
    \caption{Overview of domain tuning and task tuning}
    \label{tab:adaptabert-model}
\end{table}

\section{Evaluation Setting}
We evaluate on the task of part-of-speech tagging in the Penn Parsed Corpus of Early Modern English (\ppceme).
There is no canonical training/test split for this data, so we follow \citeauthor{moon2007part} in randomly selecting 25\% of the documents for the test set.
Details of this split are described in supplement.

\subsection{Systems}
We evaluate the following systems:
\begin{description}
\item[Frozen BERT.] This baseline applies the pretrained ``BERT-base'' contextualized embeddings, and then learns a tagger from the top-level embeddings, supervised by the PTB labeled data.
  The embeddings are from the pretrained case-sensitive BERT model, and are not adjusted during training. \newcite{peters-etal-2019-tune} refer to this as a ``feature extraction'' application of pretrained embeddings. \footnote{Note that this baseline learns only a linear final layer over the pretrained embeddings. We do not adopt weighted combination of internal contextual layers or any non-linear final layers in \newcite{peters-etal-2019-tune}.}
\item[Task-tuned BERT.] This baseline starts with pretrained BERT contextualized embeddings, and then fine-tunes them for the part-of-speech tagging task, using the PTB labeled data. This directly follows the methodology for named entity recognition proposed by \newcite{devlin2018bert} in the original BERT paper.
\item[AdaptaBERT.] Here we fine-tune the BERT contextualized embedding first on unlabeled data as described in \autoref{sec:model-fine-tune}, and then on source domain labeled data. The target domain data is the unlabeled \ppceme training set.
\item[Fine-tuned BERT.] In \emph{supervised} learning, we fine-tune the BERT contextualized embeddings on the labeled \ppceme training set. Performance of this method should be viewed as an upper bound, because large-scale labeled data is not available in many domains of interest.
\end{description}
All BERT systems use the pretrained models from Google and the PyTorch implementation from huggingface.\footnote{Models retrieved from \url{https://github.com/google-research/bert} on March 14, 2019; implementation retrieved from \url{https://github.com/huggingface/pytorch-pretrained-bert} also on March 14, 2019.} Fine-tuning was performed using one NVIDIA GeForce RTX 2080 TI GPU. Domain-adaptive fine-tuning took 12 hours, and task tuning took an additional 30 minutes.

\subsection{Previous results}
We compare the above systems against prior published results from three feature-based taggers:
\begin{description}
\item[SVM] A support vector machine baseline tagger, using the surface features described by \newcite{yang2015unsupervised}.
\item[FEMA] This is a feature-based unsupervised domain adaptation method for structured prediction~\cite{yang2015unsupervised}, which has the best reported performance on tagging the \ppceme. Unlike AdaptaBERT, the reported results for this system are based on feature induction from the entire \ppceme, including the (unlabeled) test set.
\end{description}

\subsection{Tagset mappings}
\label{sec:eval-tagset}
Because we focus on unsupervised domain adaptation, it is not possible to produce tags in the historical English (PPCHE) tagset, which is not encountered at training time.
Following \newcite{moon2007part}, we evaluate on a coarsened version of the PTB tagset, using the first letter of each tag (e.g., \ptbtag{vbd} $\to$ \ptbtag{v}).
For comparison with \newcite{yang2016part}, we also report results on the full PTB tagset.
In these evaluations, the ground truth is produced by applying the mapping of \citeauthor{moon2007part} to the \ppceme tags.

\begin{table*}
  \centering
  \begin{tabular}{@{}lllll@{}}
    \toprule
    & \multicolumn{3}{l}{Early Modern English} & PTB\\
    \cmidrule(r){2-4}
    \cmidrule(r){5-5}
    System & Accuracy & In-vocab & Out-of-vocab & Accuracy\\
    \midrule
    \textit{Unsupervised domain adaptation}\\
      1. Frozen BERT & 77.7 & 83.7 & 61.0 & 91.4 \\
      2. Task-tuned BERT & 85.3 & 90.4 & 71.1 & 98.2 \\
      3. AdaptaBERT 
      (this work) & \textbf{89.8} & \textbf{90.8} & \textbf{86.8} & \textbf{98.2} \\
      [6pt]
      \textit{Supervised in-domain training}\\
      4. Fine-tuned BERT & 98.8 & 99.0$\dagger$ & 93.2$\dagger$ & 92.4 \\
      \bottomrule
    \end{tabular}
    \caption{Tagging accuracy on \ppceme, using the coarse-grained tagset. The unsupervised systems never see labeled data in the target domain of Early Modern English.
      $\dagger$ in line 4, ``in-vocab'' and ``out-of-vocab'' refer to the \ppceme training set vocabulary; for lines 1-3, this refers the PTB training set.
    }
    \label{tab:ppceme-results-coarse-tags}
\end{table*}

\section{Results}
\label{sec:results}
Fine-tuning to the task and domain each yield significant improvements in performance over the Frozen BERT baseline (\autoref{tab:ppceme-results-coarse-tags}, line 1).
Task-tuning improves accuracy by 7.6\% on the coarse-grained tagset (line 2), and domain-adaptive fine-tuning yields a further 4.5\% in accuracy (line 3).
AdaptaBERT's performance gains are almost entirely due to the improvement on out-of-vocabulary terms, as discussed below.

The rightmost column of the table shows performance on the Penn Treebank test set. Interestingly, domain-adaptive fine-tuning has no impact on the performance on the original tagging task.
This shows that adapting the pretrained BERT embeddings to the target domain does not make them less useful for tagging in the source domain, as long as task-tuning is performed after domain-adaptive tuning.
In contrast, \emph{supervised} fine-tuning in the target domain causes performance on the PTB to decrease significantly, as shown in line 4.
This can be viewed as a form of \emph{catastrophic forgetting}~\cite{mccloskey1989catastrophic}.

As a secondary evaluation, we measure performance on the full PTB tagset in \autoref{tab:ppceme-results-ptb-tags}, thereby enabling direct comparison with prior work~\cite{yang2016part}.
AdaptaBERT outperforms task-tuned BERT by 3.9\%, again due to improvements on OOV terms. 
Task-tuned BERT is on par with the best previous unsupervised domain adaptation result (FEMA), showing the power of contextualized word embeddings, even across disparate domains.
Note also that FEMA's representation was trained on the entire \ppceme, including the unlabeled test set, while the AdaptaBERT model uses the test set only for evaluation.

\subsection{Out-of-vocabulary terms}
We define out-of-vocabulary terms as those that are not present in the PTB training set.
Of the out-of-PTB-vocabulary words in the \ppceme test set, 52.7\% of the types and 82.2\% of the tokens appear in the \ppceme training set.
This enables domain-adaptive fine-tuning to produce better representations for these terms, making it possible to tag them correctly.
Indeed, AdaptaBERT's gains come almost entirely from these terms, with an improvement in OOV accuracy of 25.8\% over the frozen BERT baseline and 15.7\% over task-tuned BERT.
Similarly, on the full PTB tagset, AdaptaBERT attains an improvement in OOV accuracy of 11.3\% over FEMA, which was the previous state-of-the-art.

\begin{table*}
    \centering
    \begin{tabular}{@{}llll@{}}
    \toprule
    System & Accuracy & In-vocab & Out-of-vocab \\
      \midrule
      1. SVM & 74.2 & 81.7 & 49.9 \\
      2. FEMA~\cite{yang2016part} & 77.9 & 82.3 & 63.2\\
      3. Task-tuned BERT & 78.4 & 84.4 & 58.4\\
      4. AdaptaBERT
      (this work) & \textbf{82.3} & \textbf{84.7} & \textbf{74.5} \\
      \bottomrule
    \end{tabular}
    \caption{Tagging accuracy on \ppceme, using the full PTB tagset to compare with \newcite{yang2016part}. }
    \label{tab:ppceme-results-ptb-tags}
\end{table*}

\subsection{Errors on in-vocabulary terms}
The final two lines of \autoref{tab:ppceme-results-coarse-tags} indicate that there remains a significant gap between AdaptaBERT and the performance of taggers trained with in-domain data: fine-tuning BERT on the \ppceme training set reduces the error rate to 1.2\%.
This improvement is largely attributable to \emph{in-vocabulary} terms: although fine-tuned BERT does better than AdaptaBERT on both IV and OOV terms, the IV terms are far more frequent.
The most frequent errors for AdaptaBERT are on tags for \example{to} (5337), \example{all} (2054), and \example{that} (1792).
The OOV term with the largest number of errors is \example{al} (306), which is a shortening of \example{all}.

These errors on in-vocabulary terms can be explained by inconsistencies in annotation across the two domains: 
\begin{itemize}
\item In the \ppceme, \example{to} may be tagged as either infinitival (\pchetag{to}, e.g., \example{I am going to study}) or as a preposition (\pchetag{p}, e.g., \example{I am going to Italy}). However, in the PTB, \example{to} is tagged exclusively as \ptbtag{to}, which is a special tag reserved for this word.\footnote{In this discussion, \pchetag{sans-serif} is used for \ppceme tags, and \ptbtag{small caps} is used for the PTB tags.} Unsupervised domain adaptation generally fails predict the preposition tag for \example{to} when it appears in the \ppceme.
\item In the \ppceme, \example{all} is often tagged as a quantifier (\pchetag{q}), which is mapped to adjective (\ptbtag{jj}) in the PTB.
However, in the PTB, these cases are tagged as determiners (\ptbtag{dt}), and as a result, the domain adaptation systems always tag \example{all} as a determiner.
\item In the PTB, the word \example{that} is sometimes tagged as a wh-determiner (\ptbtag{wdt}), in cases such as \example{symptoms that showed up decades later}.
In the \ppceme, all such cases are tagged as complementizers (\pchetag{c}), and this tag is then mapped to the preposition \ptbtag{in}.
AdaptaBERT often incorrectly tags \example{that} as \ptbtag{wdt}, when \ptbtag{in} would be correct.
\end{itemize}
These examples point to the inherent limitations of unsupervised domain adaptation when there are inconsistencies in the annotation protocol.

\section{Social Media Microblogs}
As an additional evaluation, we consider social media microblogs, of which Twitter is the best known example in English.
Twitter poses some of the same challenges as historical text: orthographic variation leads to a substantial mismatch in vocabulary between the target domain and source training documents such as Wikipedia~\cite{baldwin2013noisy,eisenstein2013bad}.
We hypothesize that domain-adaptive fine-tuning can help to produce better contextualized word embeddings for microblog text.

Our evaluation is focused on the problem of identifying named entity spans in Tweets, which was the shared task of the 2016 Workshop on Noisy User Text \cite[\wnut;][]{strauss2016results}.
In the shared task, systems were given access to labeled data in the target domain;
in contrast, we are interested to measure whether it is possible to perform this task without access to such data.
As training data, we use the canonical \conll 2003 shared task dataset, in which named entity spans were annotated on a corpus of newstext~\cite{sang2003introduction}.
Because the entity types are different in the \wnut and \conll corpora, we focus on the \emph{segmentation task} of identifying named entity spans.
Participants in the 2016 \wnut shared task competed on this metric, enabling a direct comparison with the performance of these supervised systems.

\begin{table*}
  \centering
  \begin{tabular}{@{}llllll@{}}
    \toprule
    & \multicolumn{4}{l}{\textbf{\wnut}} & \conll\\
    \cmidrule(r){2-5}
    \cmidrule(r){6-6}
    System & domain adaptation data & Precision & Recall & \textbf{F1} & F1\\
    \midrule
    \textit{Unsupervised domain adaptation}\\
      1. Task-tuned BERT & n/a & 50.9 & 66.6 & 57.7 & \textbf{97.8}\\
      2. AdaptaBERT & \wnut training & 52.8 & 66.7 & 58.9 & 97.6\\
      3. AdaptaBERT & + 1M tweets & 53.6 & 68.3 & 60.0 & 97.8\\
      4. AdaptaBERT & \wnut train+test & 57.7 & 68.9 & \textbf{62.8} & 97.8\\[6pt]
      \textit{Supervised in-domain training}\\
      5. Fine-tuned BERT & n/a & 66.3 & 62.3 & 64.3 & 80.9\\
      6. \newcite{limsopatham-collier-2016-bidirectional} & n/a & 73.5 & 59.7 & \textbf{65.9} &  \\
      \bottomrule
    \end{tabular}
    \caption{Named entity segmentation performance on the \wnut test set and \conll test set A. \newcite{limsopatham-collier-2016-bidirectional} had the winning system at the 2016 \wnut shared task. Their results are reprinted from their paper, which did not report performance on the \conll dataset.
    }
    \label{tab:ner}
\end{table*}

Results are shown in \autoref{tab:ner}. A baseline system using task-tuned BERT (trained on the \conll labeled data) achieves an F1 of 57.7 (line 1). This outperforms six of the ten submissions to the \wnut shared task, even though these systems are trained on in-domain data. AdaptaBERT yields marginal improvements when domain-adaptive fine-tuning is performed on the \wnut training set (line 2); expanding the target domain data with an additional million unlabeled tweets yields a 2.3\% improvement over the BERT baseline (line 3). Performance improves considerably when the domain-adaptive fine-tuning is performed on the combined \wnut training and test sets (line 4).

Test set adaptation is controversial: it is not realistic for deployed systems that are expected to perform well on unseen instances without retraining, but it may be applicable in scenarios in which we desire good performance on a predefined set of documents.
The \wnut test set was constructed by searching for tweets in a narrow time window on two specific topics: shootings and cybersecurity events.
It is therefore unsurprising that test set adaptation yields significant improvements, since it can yield useful representations of the names of the relevant entities, which might not appear in a random sample of tweets.
This is a plausible approach for researchers who are interested in finding the key entities participating in such events in an pre-selected corpus of text.

When BERT is fine-tuned on labeled data in the target domain, the performance of the resulting tagger improves to 64.3\% (line 5). This would have achieved second place in the 2016 \wnut shared task. The state-of-the-art system makes use of character-level information that is not available to our models~\cite{limsopatham-collier-2016-bidirectional}. As with the evaluation on Early Modern English, we find that domain-adaptive fine-tuning does not impair performance on the source domain (\conll), but supervised in-domain training increases the source domain error rate dramatically.


\section{Related Work}
\label{sec:rw}

\paragraph{Adaptation in neural sequence labeling.}
Most prior work on adapting neural networks for NLP has focused on supervised domain adaptation, in which a labeled data is available in the target domain~\cite{mou2016transferable}. RNN-based models for sequence labeling can be adapted across domains by manipulating the input or output layers individually~\cite[e.g.,][]{yang2017transfer} or simultaneously~\cite{lin2018neural}. 
Unlike these approaches, we tackle unsupervised domain adaptation, which assumes only unlabeled instances in the target domain. In this setting, prior work has focused on \emph{domain-adversarial} objectives, which construct an auxiliary loss based on the capability of an adversary to learn to distinguish the domains based on a shared encoding of the input~\cite{ganin2016domain,purushotham2016variational}. However, adversarial methods require balancing between at least two and as many as six different objectives~\cite{kim-etal-2017-adversarial}, which can lead to instability~\cite{arjovsky2017wasserstein} unless the objectives are carefully balanced~\cite{alam-etal-2018-domain}. 

In addition to supervised and unsupervised domain adaptation, there are ``distantly supervised" methods that construct noisy target domain instances, e.g., by using a bilingual dictionary~\cite{fang2017model}.
Normalization dictionaries exist for Early Modern English~\cite{baron2008vard2} and social media~\cite{han2012automatically}, but we leave their application to distant supervision for future work.

Finally, language modeling objectives have previously been used for domain adaptation of text classifiers~\citep{ziser2018pivot}, but this prior work has focused on representation learning from scratch, rather than adaptation of a pretrained contextualized embedding model. Our work shows that models that are pretrained on large-scale data yield very strong performance even when applied out-of-domain, making them a natural starting point for further adaptation.

\paragraph{Semi-supervised learning.}
Universal Language Model Fine-tuning (ULMFiT) also involves fine-tuning on a language modeling task on the target text~\cite{howard2018universal}, but the goal is semi-supervised learning: prior work shows that accurate text classification can be achieved with fewer labeled examples, but does not consider the issue of domain shift.
ULMFiT involves a training regime in which the layers of the embedding model are gradually ``unfrozen'' during task-tuning, to avoid catastrophic forgetting.
We do not employ this approach, nor did we experiment with ULMFiT's elaborate set of learning rate schedules.
Finally, contemporaneous unpublished work has applied the BERT objective to target domain data, but like ULMFiT, this work focuses on semi-supervised classification rather than cross-domain sequence labeling~\cite{Xie2019UnsupervisedDA}.

\paragraph{Tagging historical texts.} \newcite{moon2007part} approach part-of-speech tagging of Early Modern English by projecting annotations from labeled out-of-domain data to unlabeled in-domain data.
The general problem of adapting part-of-speech tagging over time was considered by \newcite{yang2015unsupervised}.
Their approach projected source (contemporary) and target (historical) training instances into a shared space, by examining the co-occurrence of hand-crafted features.
This was shown to significantly reduce the transfer loss in Portuguese, and later in English~\cite{yang2016part}.
However, this approach relies on hand-crafted features, and does not benefit from contemporary neural pretraining architectures.
We show that pretrained contextualized embeddings yield significant improvements over feature-based methods.

\paragraph{Normalization.}
Another approach to historical texts and social media is spelling normalization~\cite[e.g.,][]{baron2008vard2,han2012automatically}, which has been shown to offer improvements in tagging historical texts~\cite{robertson2018evaluating}.
In Early Modern English, \newcite{yang2016part} found that domain adaptation and normalization are complementary.
In this paper, we have shown that domain-adaptive fine-tuning (and wordpiece segmentation) significantly improves the OOV tagging accuracy from FEMA, so future research must explore whether normalization is still necessary for state-of-the-art tagging of historical texts.

\paragraph{Domain-specific pretraining.}
Given sufficient unlabeled data in the target domain, the simplest approach may be to retrain a BERT-like model from scratch. 
BioBERT is an application of BERT to the biomedical domain, which was achieved by pretraining on more than 10 billion tokens of biomedical abstracts and full-text articles from PubMed~\cite{lee2019biobert};
SciBERT is a similar approach for scientific texts~\cite{beltagy2019scibert}.
Data on this scale is not available in Early Modern English, but retraining might be applicable to social media text, assuming the user has access to both large-scale data and sufficient computing power.
The mismatch between pretrained contextualized embeddings and technical NER corpora was explored in recent work by \newcite{dai2019using}.

\section{Conclusion}
This paper demonstrates the applicability of contextualized word embeddings to two difficult unsupervised domain adaptation tasks.
On the task of adaptation to historical text, BERT works relatively well out-of-the-box, yielding equivalent performance to the best prior unsupervised domain adaptation approach.
Domain-adaptive fine tuning on unlabeled target domain data yields significant further improvements, especially on OOV terms.
On the task of adaptation to contemporary social media, a straightforward application of BERT yields competitive results, and domain-adaptive fine tuning again offers improvements.


A potentially interesting side note is that while supervised fine-tuning in the target domain results in \emph{catastrophic forgetting} of the source domain, unsupervised target domain tuning does not.
This suggests the intriguing possibility of training a single contextualized embedding model that works well across a wide range of domains, genres, and writing styles.
However, further investigation is necessary to determine whether this finding is dependent on specific details of the source and target domains in our experiments, or whether it is a general difference between unsupervised domain tuning and supervised fine-tuning.
We are also interested to more thoroughly explore how to combine domain-adaptive and task-specific fine-tuning within the framework of continual learning~\cite{yogatama2019learning}, with the goal of balancing between these apparently conflicting objectives.

\paragraph{Acknowledgments.}
Thanks to the anonymous reviewers and to Ross Girshick, Omer Levy, Michael Lewis, Yuval Pinter, Luke Zettlemoyer, and the Georgia Tech Computational Linguistics Lab for helpful discussions of this work.
The research was supported by the National Science Foundation under award RI-1452443.

\bibliography{cite-strings,cites,cite-definitions}
\bibliographystyle{aclnatbib}
\newpage

\appendix
\section{Training/test split}
\label{app:eval}
No canonical training/test split exists for the \ppceme.
We randomly select 25\% of the documents for the test set, shown in \autoref{tab:training-test} and \autoref{tab:test-doc-list}.
The remaining 75\% of documents are used for domain-adaptive fine-tuning in the AdaptaBERT results, and for training in the supervised fine-tuning results.
For the PTB, we use sections 0-18 for task-specific fine-tuning in the AdaptaBERT results and report secondary performance on sections 19-21.

\begin{table}[h]
  \centering
  \begin{tabular}{c c c}
    \toprule
    & \# documents & \# tokens \\
    \midrule
    Train: & 333 & 1473103\\
    Test: & 115 & 488054\\
    \bottomrule
  \end{tabular}
  \caption{Statistics of the training and test set from \ppceme used in our experiments.
  }
  \label{tab:training-test}
\end{table}

\begin{table*}[h]
  \centering
  \begin{tabular}{c c c}
    \toprule
alhatton-e3-h.pos & armin-e2-p1.pos & aungier-e3-h.pos \\
aungier-e3-p2.pos & authold-e2-h.pos & bacon-e2-h.pos \\
bacon-e2-p1.pos & blundev-e2-h.pos & boethco-e1-p1.pos \\
boethco-e1-p2.pos & boethpr-e3-p2.pos & burnetcha-e3-p1.pos \\
burnetroc-e3-h.pos & burnetroc-e3-p1.pos & chaplain-e1-p2.pos \\
clowesobs-e2-p2.pos & cromwell-e1-p1.pos & cromwell-e1-p2.pos \\
delapole-e1-p1.pos & deloney-e2-p1.pos & drummond-e3-p1.pos \\
edmondes-e2-h.pos & edmondes-e2-p1.pos & edward-e1-h.pos \\
edward-e1-p1.pos & eliz-1590-e2-p2.pos & elyot-e1-p1.pos \\
eoxinden-1650-e3-p1.pos & fabyan-e1-p1.pos & fabyan-e1-p2.pos \\
farquhar-e3-h.pos & farquhar-e3-p1.pos & fhatton-e3-h.pos \\
fiennes-e3-p2.pos & fisher-e1-h.pos & forman-diary-e2-p2.pos \\
fryer-e3-p1.pos & gascoigne-1510-e1-p1.pos & gawdy-e2-h.pos \\
gawdy-e2-p1.pos & gawdy-e2-p2.pos & gifford-e2-p1.pos \\
grey-e1-p1.pos & harley-e2-h.pos & harley-e2-p1.pos \\
harleyedw-e2-p2.pos & harman-e1-h.pos & henry-1520-e1-h.pos \\
hooker-b-e2-h.pos & hoole-e3-p2.pos & hoxinden-1640-e3-p1.pos \\
interview-e1-p2.pos & jetaylor-e3-h.pos & jetaylor-e3-p1.pos \\
jopinney-e3-p1.pos & jotaylor-e2-p1.pos & joxinden-e2-p2.pos \\
jubarring-e2-p1.pos & knyvett-1630-e2-p2.pos & koxinden-e2-p1.pos \\
kpaston-e2-h.pos & kpaston-e2-p1.pos & kscrope-1530-e1-h.pos \\
leland-e1-h.pos & leland-e1-p2.pos & lisle-e3-p1.pos \\
madox-e2-h.pos & marches-e1-p1.pos & markham-e2-p2.pos \\
masham-e2-p1.pos & masham-e2-p2.pos & memo-e3-p2.pos \\
milton-e3-h.pos & mroper-e1-p1.pos & mroper-e1-p2.pos \\
nhadd-1700-e3-h.pos & nhadd-1700-e3-p1.pos & penny-e3-p2.pos \\
pepys-e3-p1.pos & perrott-e2-p1.pos & pettit-e2-h.pos \\
pettit-e2-p1.pos & proposals-e3-p2.pos & rcecil-e2-p1.pos \\
record-e1-h.pos & record-e1-p1.pos & record-e1-p2.pos \\
rhaddjr-e3-h.pos & rhaddsr-1670-e3-p2.pos & rhaddsr-1700-e3-h.pos \\
rhaddsr-1710-e3-p2.pos & roper-e1-h.pos & roxinden-1620-e2-h.pos \\
rplumpt-e1-p1.pos & rplumpt2-e1-p2.pos & shakesp-e2-h.pos \\
shakesp-e2-p1.pos & somers-e3-h.pos & stat-1540-e1-p1.pos \\
stat-1570-e2-p1.pos & stat-1580-e2-p2.pos & stat-1600-e2-h.pos \\
stat-1620-e2-p2.pos & stat-1670-e3-p2.pos & stevenso-e1-h.pos \\
tillots-a-e3-h.pos & tillots-b-e3-p1.pos & turner-e1-h.pos \\
tyndold-e1-p1.pos & udall-e1-p2.pos & underhill-e1-p2.pos \\
vicary-e1-h.pos & walton-e3-p1.pos & wplumpt-1500-e1-h.pos \\
zouch-e3-p2.pos & & \\
    \bottomrule
  \end{tabular}
  \caption{Test set documents from the \ppceme used in our experiments.
  }
  \label{tab:test-doc-list}
\end{table*}

\end{document}